\documentclass{article}

\usepackage{PRIMEarxiv}

\usepackage[utf8]{inputenc} % allow utf-8 input
\usepackage[T1]{fontenc}    % use 8-bit T1 fonts
\usepackage{hyperref}       % hyperlinks
\usepackage{url}            % simple URL typesetting
\usepackage{booktabs}       % professional-quality tables
\usepackage{amsfonts}       % blackboard math symbols
\usepackage{nicefrac}       % compact symbols for 1/2, etc.
\usepackage{microtype}      % microtypography
\usepackage{lipsum}
\usepackage{fancyhdr}       % header
\usepackage{graphicx}       % graphics
\usepackage{multirow}
\usepackage{tikz}
\usepackage{pgfplots}
\usepackage{float}
\usepackage{placeins}
\usepackage{comment}
\pgfplotsset{compat=1.18}
\usetikzlibrary{positioning, arrows.meta, fit, backgrounds, shadows.blur}
\graphicspath{{media/}}     % organize your images and other figures under media/ folder
\title{Machine Learning-Driven Multimodal Spectroscopic Liquid Biopsy for Early Multicancer Detection}

\author{
  Alejandro Leonardo García Navarro\thanks{Corresponding author: agnavarr@pa.uc3m.es} \\
  Signal Processing Group \\
  Gregorio Marañón Health Research Institute \\
  Madrid, Spain \\
  \texttt{agnavarr@pa.uc3m.es} \\
  \And
  Javier Cachón Ortiz \\
  Amber Health Solutions \\
  Madrid, Spain \\
  \texttt{jacachon@amberhs.com} \\
  \And
  Javier González Colsa \\
  Amber Health Solutions \\
  Madrid, Spain \\
  \texttt{jagonzalez@amberhs.com} \\
  \And
  Samuel García Díaz \\
  Amber Health Solutions \\
  Madrid, Spain \\
  \texttt{samuel@amberhs.com} \\
  \And
  Carlos Viadero Valderrama \\
  Amber Health Solutions \\
  Madrid, Spain \\
  \texttt{carlos@amberhs.com} 
}

\begin{document}
\maketitle

\begin{abstract}

Cancer is one of the leading causes of death worldwide, making the development of rapid, minimally invasive, label-free and scalable diagnostic strategies a major challenge in modern oncology. In this context, spectroscopic liquid biopsy has emerged as a promising alternative, as it enables the holistic characterization of biochemical alterations in biological fluids. In this work, we propose a multimodal spectroscopic liquid biopsy framework for multicancer detection based on the combination of Fourier Transform Infrared (FTIR) spectroscopy, Raman spectroscopy, and Excitation–Emission Matrix (EEM) fluorescence spectroscopy together with Machine Learning (ML) methodologies. Serum samples from breast cancer patients, colorectal cancer patients, and healthy controls were analyzed through the three spectroscopic modalities. After modality-specific preprocessing, low-level data fusion (LLDF) was employed to integrate the complementary biochemical information encoded within the different spectroscopic measurements, and classification was performed using XGBoost models. Seven experimental configurations were evaluated, including the three unimodal approaches, all pairwise bimodal configurations, and the full multimodal approach of FTIR, Raman, and EEM fluorescence. The results show that although several individual modalities achieved high discrimination performance, the multimodal fusion provided the most balanced overall results, reaching a ROC-AUC of $0.997$ for breast cancer and $0.994$ for colorectal cancer, together with highly balanced sensitivity and specificity values.

\end{abstract}


\section{Introduction}

Cancer and cancer-related diseases constitute one of the leading causes of death worldwide and are currently seen as one of the major public health challenges of this century. According to recent validated global incidence estimates, around 20 million new cancer cases and nearly 10 million cancer-related deaths were reported in 2022 \cite{sung2021global}. Furthermore, current projections indicate that the number of diagnosed cases might exceed 35 million by 2050 \cite{iarc2050}, mainly due to population growth and aging. In this context, the development of new minimally invasive approaches for cancer diagnosis and monitoring has become one of the most relevant priorities in modern medicine.

In particular, early cancer detection is strongly associated with better clinical outcomes \cite{etlzioni2003early}, including higher survival rates and a greater probability of complete recovery. For many tumor types, detecting the disease at an early stage substantially increases the effectiveness of therapeutic interventions while reducing the need for highly aggressive treatments that may compromise the patient’s overall health condition. As a result, there is growing interest within the oncology research community in developing technologies capable of identifying early physiological alterations linked to cancer, even before the disease becomes clinically evident.

Nowadays, conventional diagnostic strategies mainly rely on medical imaging, histopathological analysis, and tissue biopsy procedures. Although these methods are well established and currently represent the clinical gold standard, they still show several limitations. Tissue biopsies, for instance, are inherently invasive and may lead to complications for the patient, particularly in the case of repeated sampling for monitoring. In addition, the heterogeneous nature of tumor tissues \cite{gerlinger2012intratumor} may reduce the effectiveness of sample extraction, especially at the earliest stages of disease progression. Furthermore, despite their essential role in clinical practice, imaging techniques may also show limitations in sensitivity to the weak biochemical and physiological alterations associated with the initial development of cancer. 

In this context, liquid biopsy has emerged as a minimally invasive alternative \cite{heitzer2019current} with strong potential for both cancer detection and monitoring. Unlike conventional biopsy procedures, liquid biopsy allows for the analysis of biological fluids such as blood serum or plasma, enabling scalable and repeatable diagnostic protocols. Beyond the analysis of tumor DNA or circulating tumor cells, liquid biopsy also paves the way for broader strategies based on global chemical profiling. Among these, spectroscopic techniques have attracted particular attention due to their capability to capture systemic molecular alterations associated with cancer progression.

A major advantage of these approaches is that they do not focus on a single biomarker, but instead provide a more holistic characterization of the biopsied sample. Their spectral fingerprints simultaneously contain valuable information about the structure and composition of proteins, lipids, nucleic acids, and metabolites. Consequently, spectroscopic analysis of liquid biopsy samples, particularly serum and plasma, offers significant advantages since cancer induces complex and highly interconnected biochemical alterations at multiple physiological levels. 

Among the different spectroscopic techniques proposed in recent years for the development of minimally invasive diagnostic tools, some of the most promising are Fourier Transform Infrared (FTIR) spectroscopy \cite{baker2014using}, Raman spectroscopy \cite{kong2015raman}, and Excitation–Emission Matrix (EEM) fluorescence spectroscopy \cite{croce2014fluorescence}. FTIR spectroscopy is based on the interaction between infrared radiation and the vibrational modes of molecules, providing highly sensitive biochemical fingerprints associated with proteins, lipids, nucleic acids, and carbohydrates. Raman spectroscopy, in contrast, relies on the inelastic scattering of light to obtain complementary vibrational information with high chemical specificity and minimal sample preparation requirements. Finally, EEM fluorescence spectroscopy combines multiple excitation and emission wavelengths to generate bidimensional spectral maps that are highly sensitive to variations in endogenous fluorophores present in biological samples. Together, these techniques enable the detection of slight biochemical and metabolic alterations associated with pathological processes, including cancer development and progression.

However, the complexity and high dimensionality of spectroscopic datasets make advanced computational approaches essential for extracting clinically relevant information. In recent years, Machine Learning (ML) techniques have demonstrated remarkable potential for identifying subtle spectral patterns associated with pathological states \cite{cui2018machine, shen2020deep}, enabling the development of automated diagnostic models with high sensitivity and specificity. In this context, several studies have already demonstrated the strong potential of these combined approaches for cancer detection. In the case of FTIR spectroscopy, blood-based infrared liquid biopsy studies have reported sensitivities as high as 96\% for brain cancer detection \cite{hands2014brain}, while large-scale multicancer analyses achieved Stage I cancer detection rates of approximately 64\% at 99\% specificity. Raman-based approaches have also shown excellent diagnostic capabilities. For example, Raman spectroscopy applied to ovarian cancer screening \cite{paraskevaidi2018differentiating} achieved sensitivities and specificities of 93\% and 97\%, respectively, whereas serum Surface-Enhanced Raman Spectroscopy (SERS) combined with machine learning has reported prostate cancer classification accuracies above 94\%. Furthermore, hybrid spectroscopic strategies \cite{riosreina2019data} combining complementary vibrational techniques such as FTIR and Raman spectroscopy have demonstrated enhanced robustness and discrimination capabilities by exploiting the complementary biochemical information provided by both modalities.

In this work, we propose a multimodal spectroscopic liquid biopsy framework aimed at optimizing ML-based cancer screening performance in terms of sensitivity and specificity. Unlike most previous studies, which typically rely on a single spectroscopic modality or on the combination of two vibrational techniques, our approach integrates FTIR, Raman, and EEM fluorescence spectroscopy through a low-level data fusion strategy. This multimodal representation is used to train XGBoost classifiers for the detection of breast and colorectal cancer against healthy controls, with the objective of exploiting the complementary biochemical information provided by each spectroscopic technique. In this way, FTIR captures global infrared molecular fingerprints, Raman provides highly specific vibrational information, and EEM fluorescence contributes additional information related to endogenous fluorophores and metabolic alterations. Therefore, the main goal of this study is not only to demonstrate the feasibility of spectroscopic liquid biopsy for cancer detection, but also to evaluate whether the fusion of complementary spectroscopic modalities can improve the robustness, sensitivity, and specificity of ML-based diagnostic models in a multicancer screening scenario. Generally, we believe that this approach may help to better understand the complementary diagnostic value of multiple spectroscopic modalities when properly combined with up-to-date ML approaches and their potential role in the development of robust, data-driven liquid biopsy platforms for multicancer screening.

\section{Materials and methods}
\label{sec:materials}

\subsection{Sample collection and clinical cohort}
Serum samples were collected from patients with breast cancer, colorectal cancer, and healthy controls under the approval of the Ethics Committee of the Principado de Asturias (CEIMPA nº 2023.077). Cancer samples and data from donors included in this study were provided by the Principado de Asturias BioBank (PT20/0161), integrated in the Spanish National Biobanks and Biomodels Network and they were processed following standard operating procedures with the appropriate approval of the Ethical and Scientific Committees. All the cancer samples were confirmed by histopathological diagnosis prior to inclusion, and control samples were recruited from voluntary blood donors at the Centro Comunitario de Sangre y Tejidos de Asturias.

Cancer serum samples were received frozen on dry ice, thawed at room temperature, and immediately aliquoted into protein LoBind tubes before being stored at $-80\,^\circ$C until further processing. Control samples were processed from whole blood collected in VACUETTE 8\,mL tubes with coagulation activator and separation gel. Tubes were left to coagulate at room temperature for 1--2\,h and subsequently centrigufed at 2000$\times g$ for 10\,min at room temperature. The serum fraction, clearly separated from the cellular fraction by the gel barrier, was aliquoted into Protein LoBind tubes and stored at $-80\,^\circ$C until analysis.

The cohort comprised 100 patients with breast cancer, 100 with colorectal cancer, and 100 healthy controls. The number of patients included in each experiment varied depending on the spectroscopic modality and multimodal configuration, as detailed in Table~\ref{tab:cohort}. Learning curve analysis (later shown in Section~\ref{sec:results}) indicates that the available cohort sizes were sufficient for stable model training.

\begin{table}[!ht]
\centering
\caption{Number of spectra available per group and spectroscopic modality. For FTIR, up to three technical replicates were acquired per patient when available. Raman and EEM comprise one spectrum per patient.}
\label{tab:cohort}
\begin{tabular}{lccc}
\toprule
\textbf{Group} & \textbf{FTIR} & \textbf{Raman} & \textbf{EEM} \\
\midrule
Breast cancer      & 100 & 98  & 98  \\
Colorectal cancer  & 100 & 97  & 98  \\
Control            & 100 & 77  & 80 \\
\midrule
\textbf{Total}     & \textbf{300} & \textbf{272} & \textbf{276} \\
\bottomrule
\end{tabular}
\end{table}

\subsection{Sample Preparation \& Data Acquisition}

\subsubsection{FTIR Spectroscopy}

For FTIR, serum aliquots were pre-frozen at $-80$ $^\circ$C. Lyophilization was performed using a Telstar Cryodos $-80$ system belonging to the University of Oviedo for approximately $24$ h, operating at a nominal condenser temperature of $-70.5$ $^\circ$C and a vacuum pressure of $\approx 0.302$\,mbar. After the process, the lyophilized serum was vacuum-sealed in airtight pouches with desiccant (silica gel) and stored in a dark, dry environment for preservation. Prior to acquisition, samples were individually homogenized using an agate mortar to ensure a uniform solid phase and deposisted directly onto the ATR crystal.

Data acquisition was performed on a Shimadzu IRXross spectrometer equipped with an attenuated total reflectance (ATR) accessory. Spectra were recorded in the $4000$–$650$ $cm^{-1}$ range with a resolution of $4$ $cm^{-1}$, averaging $32$ scans per measurement, and up to three technical replicates were performed per sample. To ensure reproducibility and eliminate environmental interference, a background spectrum was recorded before each measurement, and the system was maintained under a constant dry air purge at a pressure of 1\,bar and a flow rate of 5\,L/min.

\subsubsection{Raman Spectroscopy}

In the case of Raman, the lyophilized serum was deposited on an aluminum substrate covered by disposable aluminum foil to maintain cleanliness and minimize cross-contamination. This whole system is then positioned on a sample holder incorporated with a two-axis horizontal translation system with micrometer precision. The assembly was positioned on a BAC150 probe holder with XYZ adjustment. The best three coordinate position is then searched for every sample to optimize the Raman signal.

%Blood serum obtained from freeze-dried blood samples was analyzed using Raman spectroscopy. 

%Spectra were acquired using a 785 nm excitation laser with a maximum output power of 300 mW over a Raman shift range of 30–3358 $cm^{-1}$, 

Spectra were acquired using a portable spectrometer system consisting of an Exemplar Pro-785, a 785 nm laser with a maximum output power of 300 mW, and a BAC102E-785 fiber optic probe, covering a Raman shift range of 30–3358 $cm^{-1}$, without microscope coupling. The laser was directed through an optical fiber probe focused vertically with micrometer-scale precision, and a 785\,nm notch filter was placed on the detection side to suppress back-reflected laser light. Given the variability in fluorescence background across samples, acquisition parameters were adjusted individually for each measurement, with integration times between 3 and 10\,s and 10--15 accumulations. In cases of very high absorbance, laser power was reduced to prevent sample degradation.

\begin{figure}[htbp]
    \centering
    \includegraphics[width=0.7\textwidth]{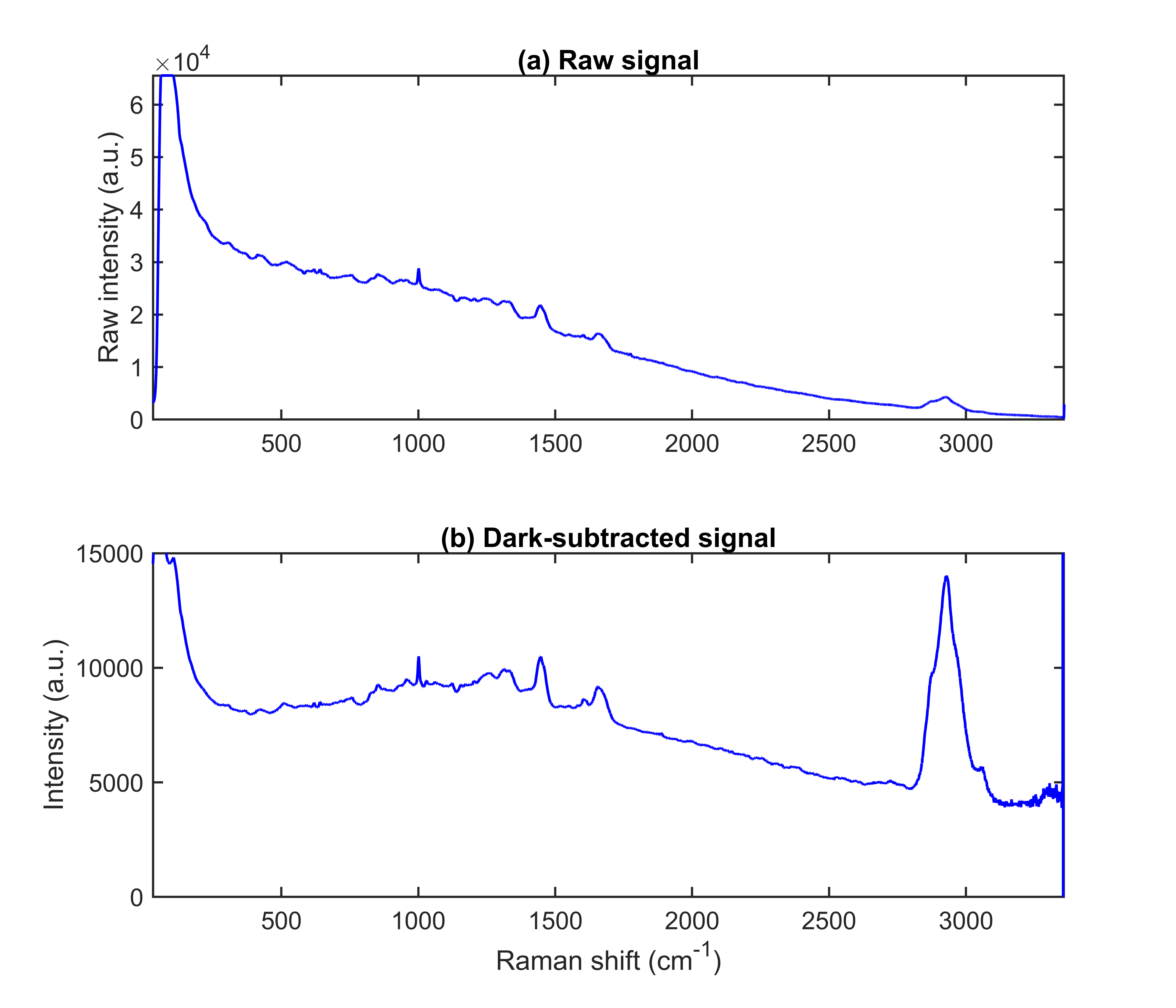}
    \caption{Raman spectra acquired from freeze-dried blood serum using a 785 nm excitation laser. (a) Raw spectrum showing a strong fluorescence background dominating the overall spectral profile, including detector saturation at low Raman shifts caused by intense fluorescence emission. Superimposed Raman bands corresponding to molecular vibrational features of the serum remain observable. (b) Dark-subtracted spectrum after background correction, highlighting the Raman contribution and improving visualization of the vibrational bands across the 30–3358 $cm^{-1}$ spectral range.
}
    \label{fig:experimental_raman}
\end{figure}

Raw spectra exhibited a strong fluorescence background superimposed on the Raman signal, with detector saturation at low Raman shifts, and etaloning artifacts were also observed in several spectra caused by internal reflections within the CCD detector. Both contributions were reduced through dark subtraction and the preprocessing steps described in Section~\ref{sec:preprocessing}. A representative raw and corrected spectrum is shown in Figure~\ref{fig:experimental_raman}.

\FloatBarrier
\subsubsection{Excitation-Emission Matrix (EEM) Fluorescence}

Finally, for EEM fluorescence, serum samples were diluted to a 1:125 ratio in phosphate-buffered saline (PBS) and placed in a high-precision quartz cuvette (Hellma Analytics) with a $10 \times 10$\,mm optical path length. A pedestal was used to align the sample volume with the instrument's optical axis, considering that the xenon arc lamp emits a beam height of 11--19\,mm.

Fluorescence analysis was performed using a Shimadzu RF-6000 
spectrofluorometer provided by Palex Healthcare Group, S.L. Excitation-emission matrices (EEMs) were recorded over an excitation range of $250$--$520$\,nm and an emission range of $270$--$750$\,nm at a scanning speed of 6000\,nm/min. Both the data interval and the bandwidth were set at 5\,nm for excitation and emission. To prevent detector saturation, a low sensitivity setting (Low Gain) was selected. In all cases, the spectrum of a control sample (PBS) measured under identical conditions was subtracted to eliminate water Raman scattering and other background interferences.

\subsection{Preprocessing pipeline}
\label{sec:preprocessing}
Raw spectroscopic signals from all three modalities require careful preprocessing before any downstream analysis, as they often contain physical artifacts such as baseline offsets, scattering effects, and high-frequency noise unrelated to the biochemical composition of the sample. This is particularly relevant in high-dimensional datasets with limited sample sizes such as those considered in this study. Each modality was therefore processed using a dedicated pipeline adapted to its acquisition characteristics and refined through experiments prior to training.

\subsubsection{FTIR Spectroscopy}

FTIR spectra are particularly sensitive to preprocessing due to baseline drifts, scattering effects, and the presence of broad overlapping absorption bands. To identify a preprocessing strategy that remained robust across the different classification scenarios, we systematically evaluated 2,880 preprocessing pipelines generated by combining alternatives at each stage of the workflow.

The evaluated preprocessing steps include replicate average, spectral region selection, baseline correction, scatter correction, smoothing, derivative computation, and intensity normalization. For dealing with replicates, both replicate averaging and no averaging were considered when multiple acquisitions were available for the sample. Different spectral regions were also explored: the full spectrum (650--4000\,cm$^{-1}$), the fingerprint region (900--1800\,cm$^{-1}$), the amide region (1500--1700\,cm$^{-1}$), the lipid region (2800--3000\,cm$^{-1}$), and the nucleic acid region (1000--1250\,cm$^{-1}$).

For baseline correction, we evaluated polynomial correction, Asymmetric Least Squares (ALS), and no correction. Scatter effects were addressed using Standard Normal Variate (SNV), which was compared against the option of applying no scatter correction. Additionally, to reduce high-frequency noise, Savitzky–Golay filtering, moving average smoothing, and no smoothing were considered.

Derivative computation included four alternatives: no derivative, first derivative, second derivative, and the concatenation of first and second derivative. Finally, intensity normalization was evaluated using area normalization, \(\ell_2\) normalization, max normalization, and no normalization.

Rather than evaluating preprocessing stages independently, each pipeline was assessed as a complete workflow, since the effect of one step often depends on the others.  The final configuration was selected using a min-max criterion: for each pipeline, the lowest ROC-AUC obtained across all cancer scenarios was identified, and the pipeline maximizing this worst-case value was chosen. This favors preprocessing strategies with stable performance across datasets over those that perform well only in specific scenarios.

\begin{figure}[htbp]
    \centering
    \includegraphics[width=\textwidth]{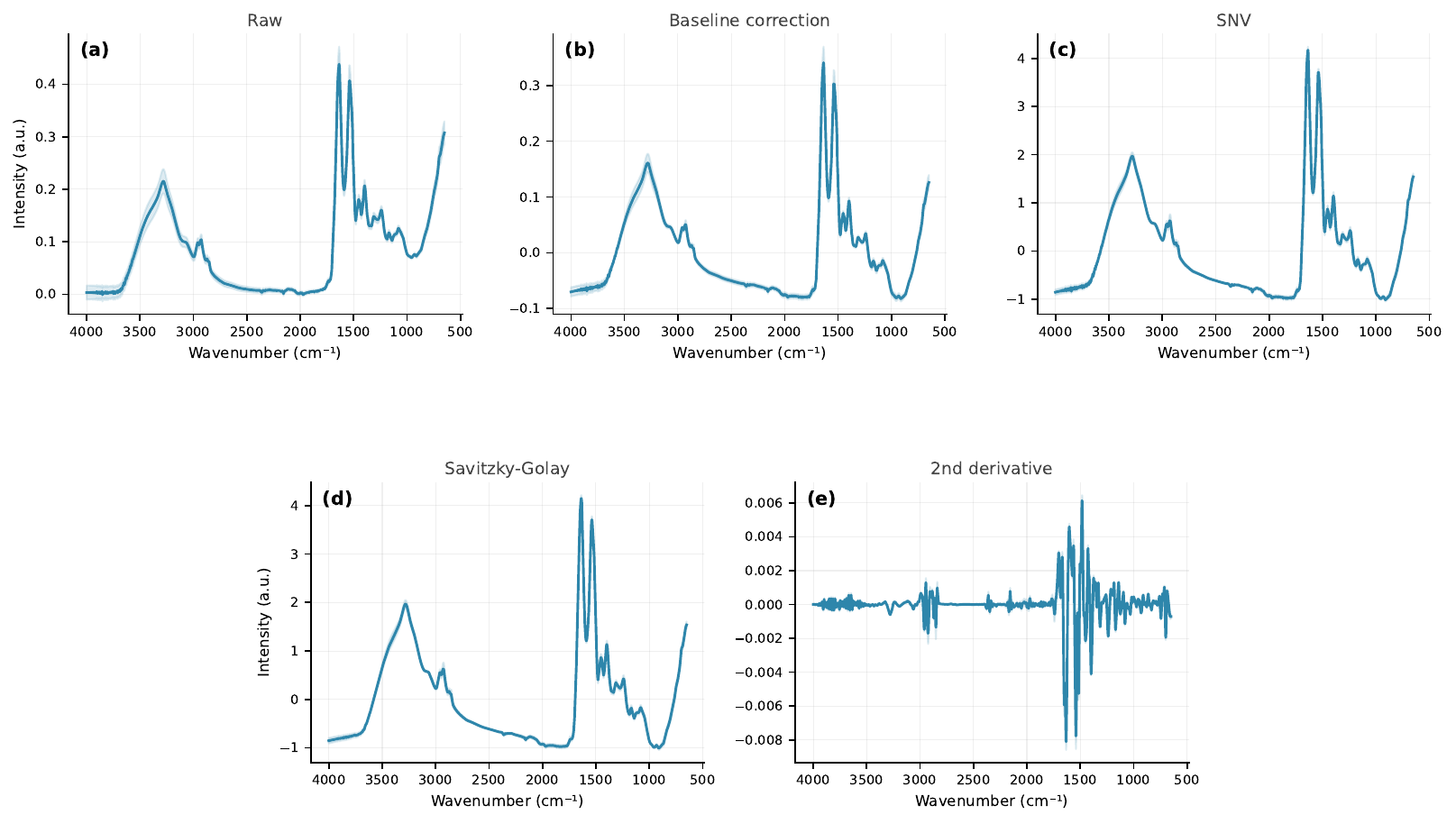}
    \caption{Sequential preprocessing pipeline applied to FTIR spectra 
    (control group, mean $\pm$ standard deviation). 
    (a) Raw spectra. 
    (b) After polynomial baseline correction. 
    (c) After SNV scatter correction. 
    (d) After Savitzky--Golay smoothing. 
    (e) After second derivative computation.}
    \label{fig:preproc_ftir}
\end{figure}

The final selected pipeline consisted of no replicate averaging, full spectral region, polynomial baseline correction, SNV scatter correction, Savitzky–Golay smoothing, second derivative computation, and no intensity normalization. The impact of these operations on the raw signals is illustrated in Figure~\ref{fig:preproc_ftir}.

\subsubsection{Raman Spectroscopy}

Compared to FTIR, Raman spectra tend to present sharper and more clearly separated peaks, which usually reduces the need for extensive preprocessing. Nevertheless, fluorescence background remains the main source of artifact and must be corrected before analysis. In this case, a fixed preprocessing pipeline was used instead of performing an exhaustive search.

\begin{figure}[htbp]
    \centering
    \includegraphics[width=\textwidth]{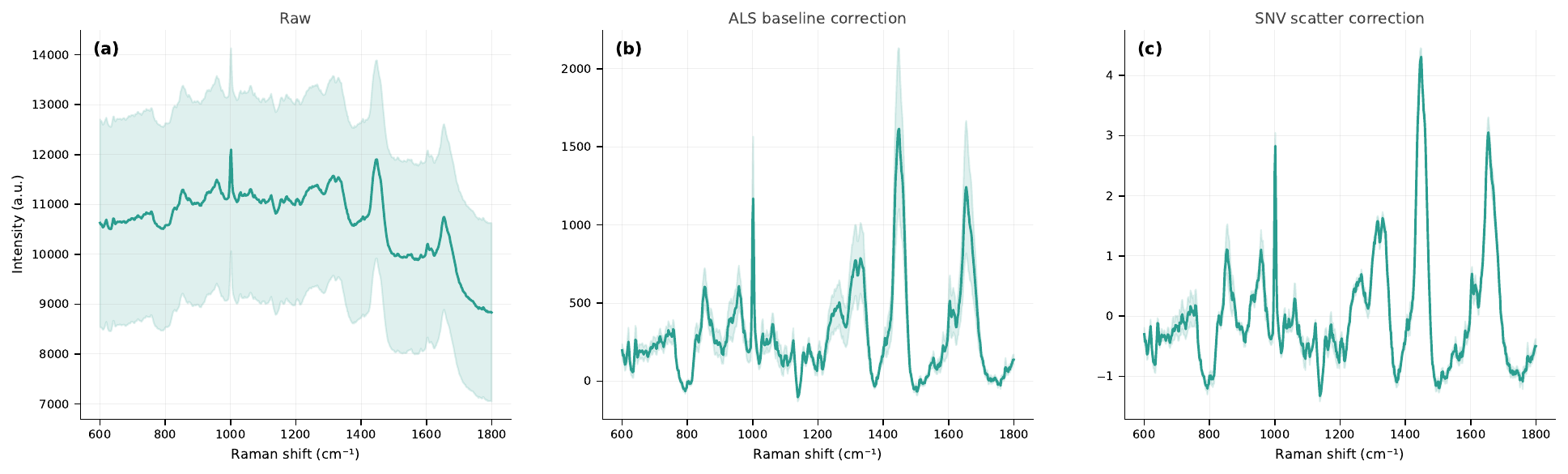}
    \caption{Sequential preprocessing pipeline applied to Raman spectra 
    (control group, mean $\pm$ standard deviation). 
    (a) Raw spectra (600--1800~$cm^{-1}$). 
    (b) After ALS baseline correction. 
    (c) After SNV scatter correction.}
    \label{fig:preproc_raman}
\end{figure}

The signal was restricted to the 600–1800 $cm^{-1}$ spectral region, which contains most of the Raman-active molecular vibrations relevant for cancer characterization~\cite{KONG2015121, Shi}. Baseline correction was performed using ALS to remove the varying fluorescence background commonly observed in this type of spectra. Finally, scatter correction was applied using SNV. The resulting preprocessed Raman spectra are shown in Figure~\ref{fig:preproc_raman}.

\subsubsection{Excitation-Emission Matrix (EEM) Fluorescence}
Unlike FTIR and Raman spectroscopy, which produce one-dimensional spectra, EEM fluorescence measurements generate two-dimensional matrices in which rows and columns correspond to excitation and emission wavelengths, respectively. This different structure required preprocessing steps specific to the modality following common steps used in the literature.

First, a phosphate-buffered saline (PBS) blank measurement acquired under the same instrumental conditions and on the same acquisition date was subtracted for each sample matrix in order to remove background fluorescence contributions from the solvent and the cuvette~\cite{3d525b947d8b481e8ddfbc36f450e4fa, 
C3AY41160E}.

To ensure physical consistency, all matrix entries where \(\lambda_{\mathrm{em}} < \lambda_{\mathrm{ex}}\) were set to zero, since emission at wavelengths shorter than the excitation wavelength is physically impossible~\cite{3d525b947d8b481e8ddfbc36f450e4fa}. Rayleigh scatter artifacts were also removed. These artifacts appear as strong signals around the diagonals where \(\lambda_{\mathrm{em}} \approx \lambda_{\mathrm{ex}}\) (first-order scatter) and \(\lambda_{\mathrm{em}} \approx 2\lambda_{\mathrm{ex}}\) (second-order scatter), and can dominate the feature space if not corrected~\cite{bahram2006}. Both scatter regions were removed by masking a symmetric window of 25nm around each diagonal. No interpolation was applied to reconstruct the masked regions, as this could introduce artificial patterns into the data.

\begin{figure}[htbp]
    \centering
    \includegraphics[width=\textwidth]{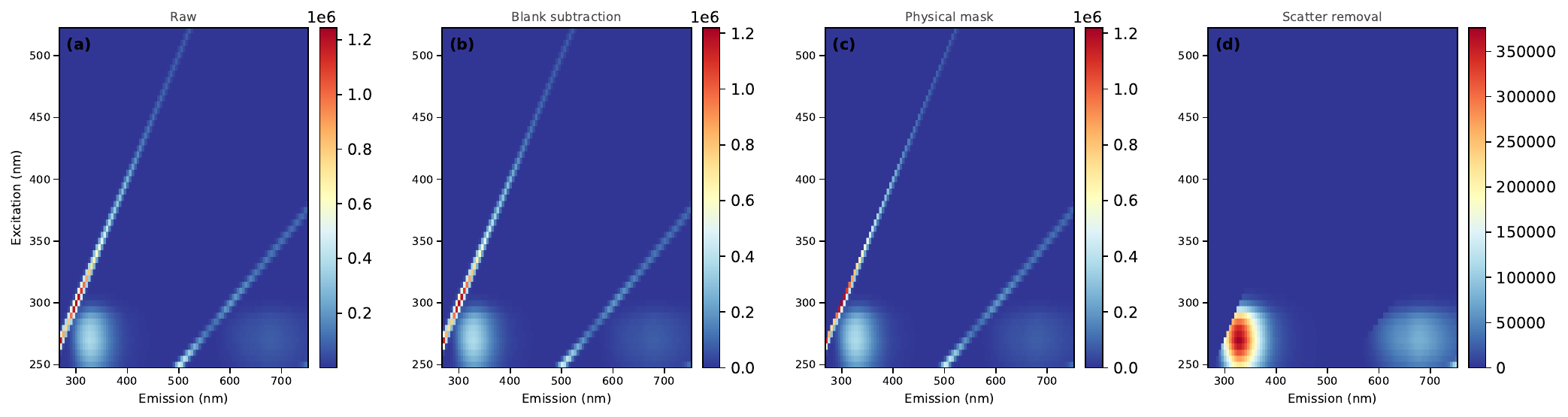}
    \caption{Sequential preprocessing pipeline applied to EEM fluorescence 
    data (representative control sample). 
    (a) Raw acquisition. 
    (b) After PBS blank subtraction. 
    (c) After physical validity masking ($\lambda_\mathrm{em} < \lambda_\mathrm{ex}$). 
    (d) After Rayleigh scatter removal (first and second order).}
    \label{fig:preproc_eem}
\end{figure}

Figure~\ref{fig:preproc_eem} shows the transformation of a representative EEM sample.

\subsection{Feature extraction and representation}
After preprocessing, each sample was represented as a fixed-length numerical vector suitable for supervised classification. No additional dimensionality reduction or learned feature extraction methods were applied, and the preprocessed signals were used directly as input features.

FTIR and Raman measurements were represented as one-dimensional spectral vectors. For EEM fluorescence, each preprocessed excitation-emission matrix was flattened row-wise into a one-dimensional vector, such that each feature corresponds to a specific excitation-emission wavelength pair.

To evaluate the contribution of each modality, three experimental configurations were considered: FTIR alone, FTIR combined with Raman, and the full combination of FTIR, Raman, and EEM fluorescence. To integrate the three modalities into a single representation, a low-level data fusion (LLDF) strategy was adopted~\cite{ijms252010936}, in which feature vectors are combined prior to classification, allowing the model to capture correlations across modalities. 

Before concatenation, two scaling steps were applied within each cross-validation fold to ensure that the fused representation is not dominated by a single modality. First, each modality $m \in \{\mathrm{FTIR, Raman, EEM}\}$ was standardized feature-wise using z-score normalization~\cite{CAMPOS2020103959, SILVESTRI2014181, MISHRA2021116206}:

\begin{equation}
    \tilde{x}_{ij}^{(m)} = \frac{x_{ij}^{(m)} - \mu_j^{(m)}}{\sigma_j^{(m)}}
    \label{eq:zscore}
\end{equation}

where $\mu_j^{(m)}$ and $\sigma_j^{(m)}$ denote the mean and standard deviation of feature $j$, estimated exclusively on the training partition and subsequently applied to the test partition.

Second, block scaling was applied to compensate for the large differences in feature dimensionality across modalities~\cite{CAMPOS2020103959, https://doi.org/10.1002/cem.713, RIOSREINA2019560}. Each standardized feature $j$ of modality $m$ was further divided by a modality-specific factor:

\begin{equation}
    \hat{x}_{ij}^{(m)} = \frac{\tilde{x}_{ij}^{(m)}}{d_m^{1/4}}
    \label{eq:blockscaling}
\end{equation}

where \(d_m\) denotes the number of features in modality \(m\). Scaling by \(d_m^{1/4}\) was used to balance the contribution of each modality while accounting for differences in dimensionality.

\begin{figure}[htbp]
    \centering
    \includegraphics[width=\textwidth]{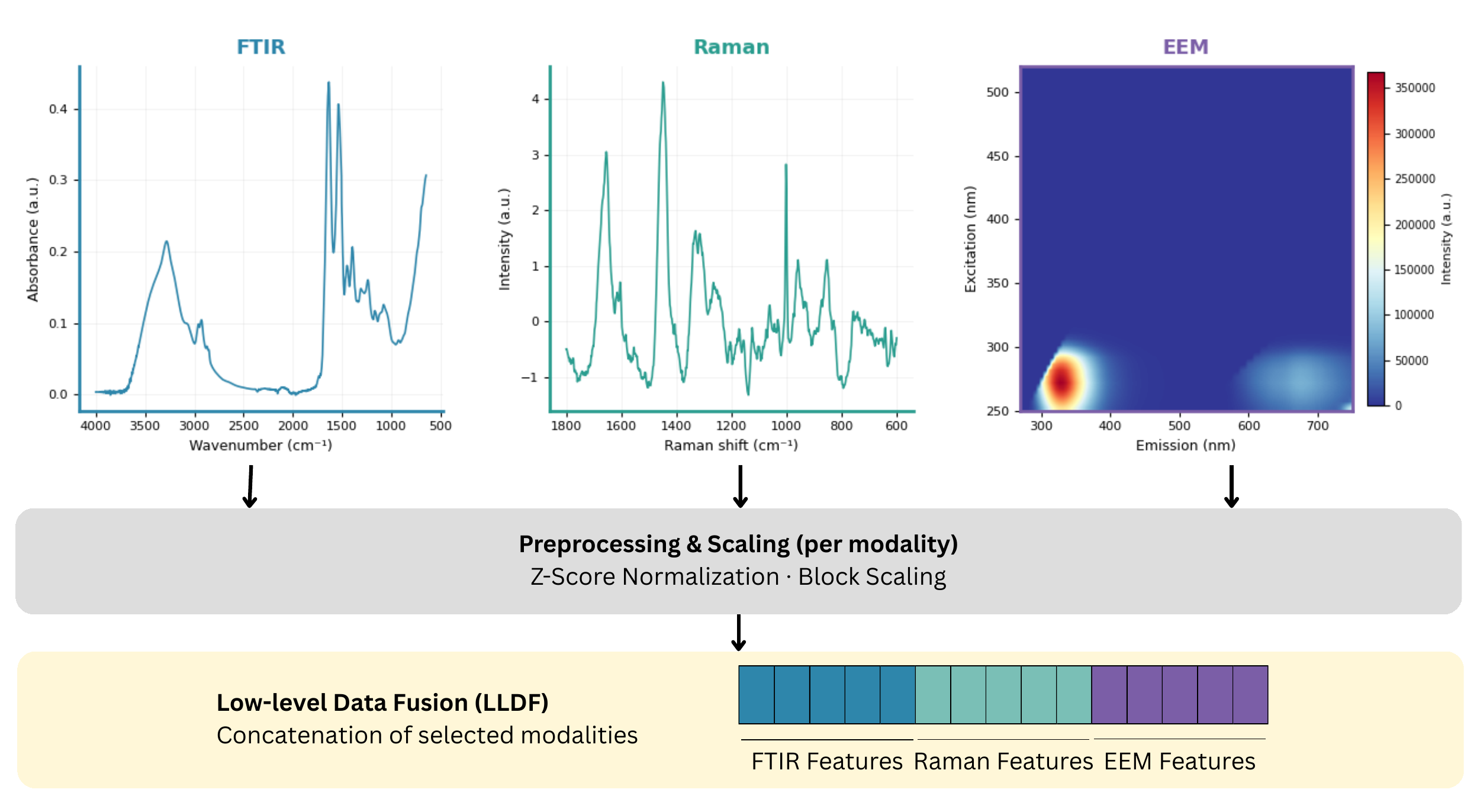}
    \caption{Low-level data fusion pipeline. Each modality vector is 
    independently standardized via z-score normalization and rescaled 
    by block scaling before concatenation into a single fused feature 
    vector.}
    \label{fig:lldf}
\end{figure}

The final feature vector for a given configuration was obtained by concatenating the corresponding modality-specific vectors, as illustrated in Figure~\ref{fig:lldf}:

\begin{equation}
    \mathbf{x} = \left[ 
        \hat{\mathbf{x}}^{(m_1)} \;\Big|\; \cdots \;\Big|\; \hat{\mathbf{x}}^{(m_k)}
    \right]
    \label{eq:fusion}
\end{equation}

where \(\{m_1, \ldots, m_k\}\) denotes the subset of modalities included in the corresponding experimental configuration. This vector served as input to the classification models described in the following section.

\subsection{Classification Framework}
\label{sec:classifier}
In this study, we addressed the diagnostic task through a series of independent binary classification experiments. For each cancer type in the cohort, a classifier was trained to distinguish between pathological and control samples using all available modality configurations. 

All classification experiments were performed using Gradient Boosted Decision Trees, specifically the XGBoost implementation. This choice was motivated by the characteristics of the dataset: high-dimensionality, limited number of samples, and complex relationships between features across modalities. 

All hyperparameters were kept at their default values throughout the study, and this is due to the exact purpose that the goal of this work was  to evaluate the impact of preprocessing and multimodal fusion strategies. Using fixed parameters helps ensure that performance differences across experiments are due to input representations rather than by model tuning.

Model evaluation was performed using stratified 10-fold cross-validation in order to preserve class proportions across folds. When replicate averaging was not applied, patient-level grouping was enforced during splitting to ensure that measurements from the same patient could not appear at the same time in training and test partitions. This prevents information leakage and avoids overfitting.

Performance was evaluated using ROC-AUC, balanced accuracy, sensitivity, and specificity. These metrics were selected to provide a threshold-independent evaluation and to reduce the effect of class imbalance, which is not adequately captured by accuracy alone.

\section{Results}
\label{sec:results}

\begin{figure}[htbp]
    \centering
    \includegraphics[width=\textwidth]{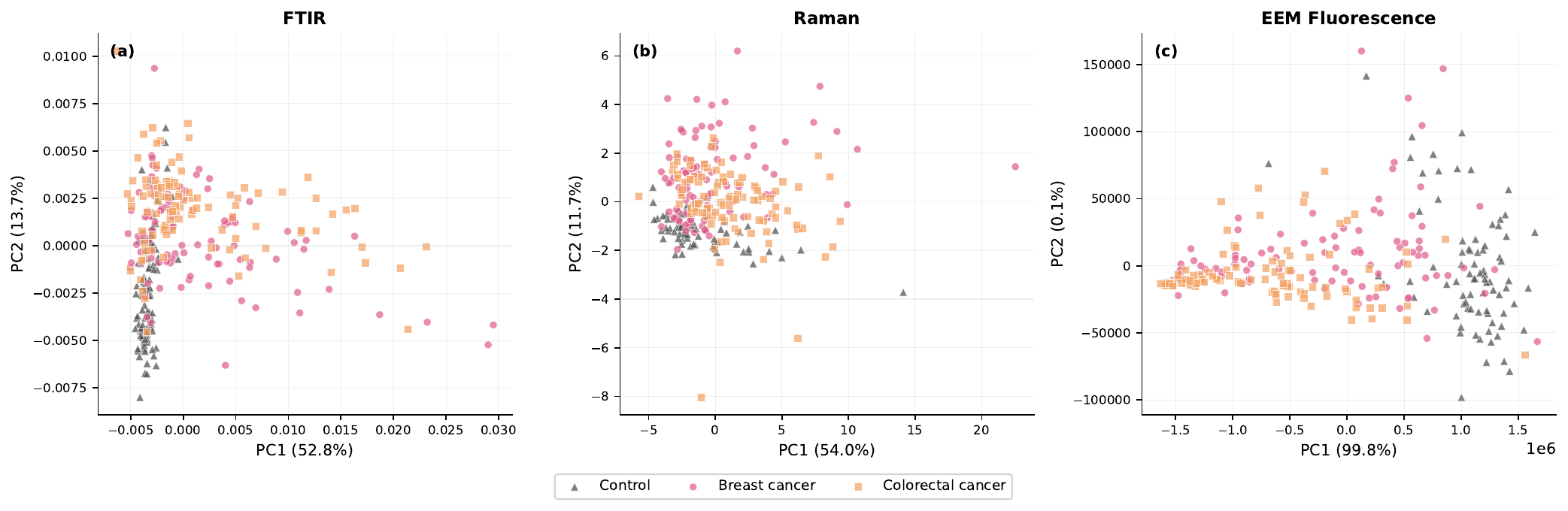}
    \caption{
    PCA projection of the preprocessed spectroscopic data for the three unimodal
    representations: (a) FTIR, (b) Raman, and (c) EEM fluorescence.
    Each point corresponds to one patient sample colored according to the clinical group.
    The first two principal components are shown together with the percentage of explained variance.
    }
    \label{fig:pca_modalities}
\end{figure}

To provide an initial overview of the spectroscopic data structure, Figure~\ref{fig:pca_modalities} illustrates the PCA projections obtained for each type of signal. While the first two principal components do not completely separate the clinical groups, different clustering tendencies can be observed across modalities, suggesting that each spectroscopic technique captures complementary biochemical information.

\begin{figure}[htbp]
    \centering
    \includegraphics[width=\textwidth]{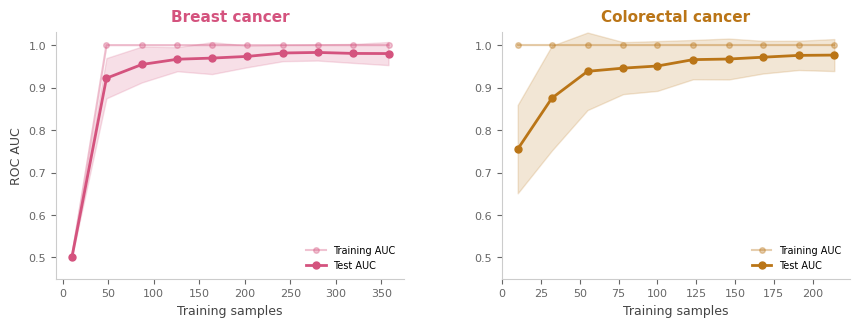}
    \caption{Learning curves for the FTIR unimodal configuration showing 
    ROC-AUC as a function of the number of training samples for breast 
    and colorectal cancer scenarios. Shaded regions represent standard 
    deviation across cross-validation folds.}
    \label{fig:learning_curve}
\end{figure}

To verify that the number of samples available per modality was sufficient  for stable model training, learning curves were computed for the FTIR  configuration as a representative case, given that it provides the largest cohort among the three modalities. Figure~\ref{fig:learning_curve} shows the ROC-AUC as a function of the number of training samples for both cancer scenarios. 

In both cases, the test AUC stabilizes rapidly, reaching near-plateau performance with fewer than 75 training samples, and remains stable as the training set grows. This suggests that the variable number of patients across modality configurations is unlikely to introduce significant performance bias, and that the available cohort sizes are sufficient for reliable model training.

Having established that the available cohort sizes are sufficient for stable model training, Table~\ref{tab:results} summarizes the classification performance obtained across all experiments. Results are reported as mean and standard deviation over the 10 cross-validation folds.

\begin{table}[h]
\centering
\caption{
Classification performance across cancer types and modality configurations.
Best values per scenario are highlighted in bold. 
$n_\mathrm{cancer}$ and $n_\mathrm{control}$ denote the number of cancer and healthy control patients included in each experiment. 
For FTIR unimodal, the model is trained at the replicate level (multiple acquisitions per patient); for all other configurations, data are aligned at the patient level, as Raman and EEM provide a single measurement per patient.
}
\label{tab:results}
\resizebox{\textwidth}{!}{%
\begin{tabular}{llcccccccc}
\toprule
\textbf{Scenario} &
\textbf{Configuration} &
\textbf{$n$} &
\textbf{$n_\mathrm{cancer}$} &
\textbf{$n_\mathrm{control}$} &
\textbf{AUC} &
\textbf{Sensitivity} &
\textbf{Specificity} &
\textbf{Bal.\ Accuracy} \\
\midrule
\multirow{7}{*}{Breast}
  & FTIR                & 200 & 100 & 100 & $0.979 \pm 0.015$ & $0.940 \pm 0.066$ & $0.909 \pm 0.070$ & $0.924 \pm 0.046$ \\
  & Raman               & 175 & 98  & 77  & $\mathbf{0.999 \pm 0.004}$ & $0.969 \pm 0.065$ & $\mathbf{0.961 \pm 0.060}$ & $0.965 \pm 0.037$ \\
  & EEM                 & 178 & 98  & 80  & $0.982 \pm 0.016$ & $0.960 \pm 0.049$ & $0.900 \pm 0.094$ & $0.930 \pm 0.052$ \\
  & FTIR + Raman        & 167 & 98  & 69  & $0.997 \pm 0.006$ & $0.980 \pm 0.040$ & $0.969 \pm 0.062$ & $\mathbf{0.975 \pm 0.032}$ \\
  & FTIR + EEM          & 175 & 98  & 77  & $0.988 \pm 0.018$ & $0.970 \pm 0.064$ & $0.923 \pm 0.115$ & $0.947 \pm 0.057$ \\
  & Raman + EEM         & 169 & 97  & 72  & $0.997 \pm 0.006$ & $0.978 \pm 0.044$ & $0.946 \pm 0.066$ & $0.962 \pm 0.049$ \\
  & FTIR + Raman + EEM  & 166 & 97  & 69  & $0.997 \pm 0.006$ & $\mathbf{0.990 \pm 0.030}$ & $0.926 \pm 0.133$ & $0.958 \pm 0.066$ \\
\midrule
\multirow{7}{*}{Colon}
  & FTIR                & 200 & 100 & 100 & $0.961 \pm 0.046$ & $0.890 \pm 0.114$ & $0.939 \pm 0.081$ & $0.914 \pm 0.060$ \\
  & Raman               & 174 & 97  & 77  & $0.983 \pm 0.026$ & $0.958 \pm 0.085$ & $0.898 \pm 0.110$ & $0.928 \pm 0.074$ \\
  & EEM                 & 178 & 98  & 80  & $0.976 \pm 0.039$ & $0.929 \pm 0.090$ & $0.963 \pm 0.057$ & $0.946 \pm 0.046$ \\
  & FTIR + Raman        & 166 & 97  & 69  & $0.987 \pm 0.020$ & $\mathbf{0.980 \pm 0.040}$ & $0.883 \pm 0.154$ & $0.932 \pm 0.072$ \\
  & FTIR + EEM          & 175 & 98  & 77  & $0.984 \pm 0.024$ & $0.959 \pm 0.067$ & $\mathbf{0.963 \pm 0.080}$ & $\mathbf{0.961 \pm 0.052}$ \\
  & Raman + EEM         & 168 & 96  & 72  & $0.990 \pm 0.011$ & $0.959 \pm 0.050$ & $0.945 \pm 0.068$ & $0.952 \pm 0.037$ \\
  & FTIR + Raman + EEM  & 165 & 96  & 69  & $\mathbf{0.994 \pm 0.019}$ & $0.959 \pm 0.050$ & $0.957 \pm 0.065$ & $0.958 \pm 0.039$ \\
\bottomrule
\end{tabular}%
}
\end{table}

The baseline models using FTIR alone, trained at the replicate level using the preprocessing pipeline selected in Section~\ref{sec:preprocessing}, achieved strong performance in both scenarios, with AUC values of $0.979$ and $0.961$ and balanced accuracies above $0.91$ for breast and colorectal cancer, respectively. The higher variability in colon sensitivity
($0.890 \pm 0.114$) suggests that FTIR alone may capture less discriminative information between tumor and control tissue in this scenario.

Raman spectroscopy alone outperformed FTIR in the breast scenario, achieving an AUC of $0.999$ with low variance, which suggests that this type of signal is discriminative for cancer type. For colorectal cancer, Raman performed better than FTIR in balanced accuracy ($0.928$ vs.\ $0.914$), with higher sensitivity and lower specificity. EEM fluorescence also showed consistent performance across both scenarios (AUC $0.982$ and $0.976$), achieving competitive balanced accuracies.

Multimodal fusion through LLDF generally improved performance compared to the unimodal baselines, although the optimal configuration varied across cancer types. In the breast cancer scenario, combining FTIR and Raman increased the AUC to $0.997$ and the balanced accuracy to $0.975$, with improvements in both sensitivity and specificity, supporting the complementary nature of the two modalities. The Raman + EEM combination also reached an AUC of $0.997$ with a balanced accuracy of $0.962$, suggesting that FTIR is not strictly necessary to achieve high discrimination in this scenario. In the colon scenario, FTIR + Raman achieved the highest sensitivity ($0.980$), although specificity became more variable across folds ($0.883 \pm 0.154$), while FTIR + EEM provided a more stable result with the highest balanced accuracy among bimodal configurations ($0.961 \pm 0.052$).

\begin{comment}
The addition of Raman spectroscopy to FTIR through LLDF led to consistent improvements in the breast scenario, where AUC increased from from $0.985$ to $0.997$ and balanced accuracy improved from $0.923$ to $0.975$. However, the improvement for colon cancer detection was more limited. While sensitivity increased to $0.980$, specificity became more variable across folds ($0.883 \pm 0.154$), leading to only a modest gain in balanced accuracy. 
\end{comment}

The integration of all three modalities produced competitive results in both scenarios. In breast cancer, FTIR + Raman + EEM maintained the high AUC of $0.997$ achieved by the best bimodal configurations while reaching the highest sensitivity ($0.990 \pm 0.030$), though at the cost of increased variability in specificity ($0.926 \pm 0.133$). In colorectal 
cancer, the trimodal configuration achieved the highest AUC ($0.994 \pm 0.019$), with sensitivity and specificity both close to $0.96$.

\begin{figure}[htbp]
    \centering
    \includegraphics[width=\textwidth]{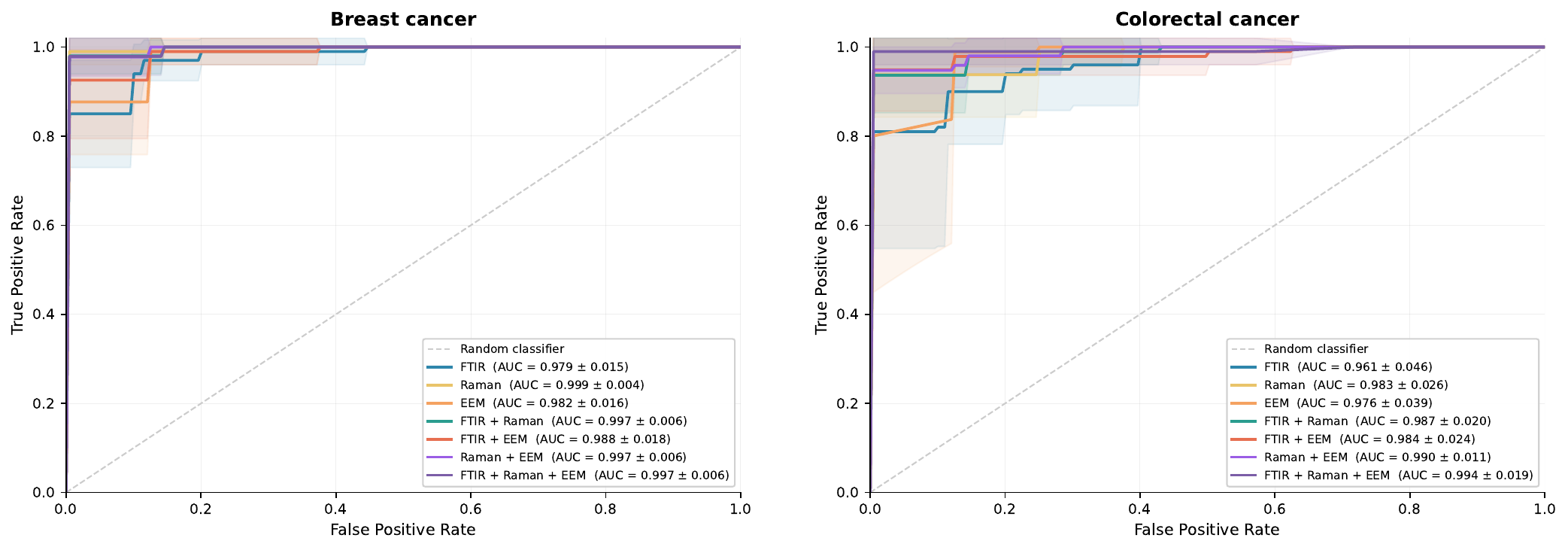}
    \caption{Mean ROC curves (± standard deviation across folds) for breast and colorectal cancer classification across all modality configurations. 
    Shaded regions represent fold-to-fold variability.}
    \label{fig:roc}
\end{figure}

The ROC curves corresponding to all modality configurations are shown in Figure~\ref{fig:roc}, further showing the classification performance achieved across both cancer scenarios.

\section{Conclusion}
\label{sec:conclusion}
In this work, we presented a multimodal spectroscopic liquid biopsy framework for multicancer detection based on the combination on FTIR, Raman, and EEM fluorescence spectroscopy together with machine learning methodologies. Through the systematic evaluation and comparison of unimodal, bimodal and trimodal configurations, the proposed framework shows that the integration of complementary spectroscopic modalities can substantially improve the robustness and balance of ML-based diagnostic models. 

\begin{comment}
Although several individual modalities already achieved strong discrimination capabilities, in particular when it comes to ROC-AUC, the full multimodal fusion consistently provided with the most balanced performance in terms of sensitivity, specificity and balanced accuracy. In the case of breast cancer, the FTIR + Raman + EEM configuration achieved a ROC-AUC of $0.997$ with a sensitivity and specificity of $0.989$ and $0.986$ respectively. On the other hand, in the case of colorectal cancer, the trimodal approach achieved the highest ROC-AUC ($0.994$) while keeping highly balanced sensitivity and specificity close to $0.96$. 

The bimodal configurations also revealed relevant complementarities between spectroscopic modalities. In the breast cancer scenario, both FTIR + Raman and Raman + EEM combinations achieved excellent discrimination performance, reaching ROC-AUC values of $0.997$ together with balanced sensitivity and specificity metrics above $0.95$. In the colorectal cancer scenario, the FTIR + EEM combination provided the highest balanced accuracy among the bimodal approaches ($0.961$), while FTIR + Raman achieved the highest sensitivity ($0.980$), although with larger variability in specificity across folds. 
\end{comment}

Although several individual modalities already achieved strong discrimination capabilities, multimodal fusion generally provided more balanced and stable performance across metrics. The contribution of each modality proved to be cancer-type-dependent, with different configurations emerging as optimal depending on the scenario.

In the breast cancer scenario, both FTIR + Raman and Raman + EEM combinations reached an AUC of $0.997$ with balanced accuracy above $0.96$, suggesting strong complementarity between vibrational and fluorescence spectroscopy. The trimodal FTIR + Raman + EEM configuration achieved the highest sensitivity ($0.990 \pm 0.030$), although at the cost of increased variability in specificity ($0.926 \pm 0.133$).

In the colorectal cancer scenario, FTIR + EEM provided the most stable bimodal result, achieving the highest balanced accuracy ($0.961 \pm 0.052$) and specificity ($0.963 \pm 0.080$) among all configurations. FTIR + Raman reached the highest sensitivity ($0.980$) but with considerably larger variability in specificity ($0.883 \pm 0.154$). The trimodal approach achieved the highest AUC ($0.994 \pm 0.019$) with sensitivity and specificity both close to $0.96$, representing the most balanced overall result in this scenario.

These results support the hypothesis that each spectroscopic modality captures complementary information associated with cancer-related alterations. Beyond the classification performance, these results suggest the potential of low-level data fusion strategies combined with ML approaches for the integration of heterogeneous spectroscopic datasets. This framework applied to liquid biopsy may constitute a promising path for the rapid, minimaly-invasive and scalable cancer screening. 

However, several limitations should be acknowledged. The cohort size remains moderate, and although cross-validation and patient-level splitting strategies were employed to minimize overfitting and information leakage risks, external validation on independent cohorts will be necessary to assess the generalization and clinical robustness of the methodology. Thus, future work will focus on larger multicentric studies, additional cancer types, and advanced multimodal fusion strategies to better discriminate spectral patterns and interpret the biochemical origin of the data.

\section*{Acknowledgments}
\label{sec:acks}
\begin{comment}
This work was supported by the Spanish Public-Private Collaboration 
Project CPP2024-011358, ``Optimización y validación de SCANHEALTH, una 
herramienta rentable de espectroscopia infrarroja con integración de IA para la detección precoz de múltiples cánceres en biopsias líquidas.''

We want to particularly acknowledge for its collaboration, the Principado de Asturias BioBank (PT20/0161), financed jointly by Servicio de Salud del Principado de Asturias, Instituto de Salud Carlos III and Fundación Bancaria Cajastur and integrated in the Spanish National Biobanks and Biomodels Netwotk.
\end{comment}

The authors thank the University of Oviedo, Palex Healthcare Group, S.L., Microbeam, S.L., and Fundación Centro Médico de Asturias for the support, facilities, and equipment provided for this study.

%Bibliography
\bibliographystyle{unsrt}  
\bibliography{references}

@article{sung2021global,
  title={Global cancer statistics 2020: GLOBOCAN estimates of incidence and mortality worldwide for 36 cancers in 185 countries},
  author={Sung, Hyuna and Ferlay, Jacques and Siegel, Rebecca L and others},
  journal={CA: A Cancer Journal for Clinicians},
  volume={71},
  number={3},
  pages={209--249},
  year={2021},
  publisher={Wiley Online Library}
}

@misc{iarc2050,
  title={Global Cancer Observatory: Cancer Tomorrow},
  author={{International Agency for Research on Cancer}},
  year={2024},
  note={https://gco.iarc.who.int/tomorrow}
}

@article{etlzioni2003early,
  title={The case for early detection},
  author={Etzioni, Ruth and Urban, Neal and Ramsey, Scott and others},
  journal={Nature Reviews Cancer},
  volume={3},
  number={4},
  pages={243--252},
  year={2003}
}

@article{gerlinger2012intratumor,
  title={Intratumor heterogeneity and branched evolution revealed by multiregion sequencing},
  author={Gerlinger, Marco and Rowan, Andrew J and Horswell, Stuart and others},
  journal={New England Journal of Medicine},
  volume={366},
  number={10},
  pages={883--892},
  year={2012}
}

@article{heitzer2019current,
  title={Current and future perspectives of liquid biopsies in genomics-driven oncology},
  author={Heitzer, Ellen and Haque, Imran S and Roberts, Christine ES and Speicher, Michael R},
  journal={Nature Reviews Genetics},
  volume={20},
  number={2},
  pages={71--88},
  year={2019}
}

@article{baker2014using,
  title={Using Fourier transform IR spectroscopy to analyze biological materials},
  author={Baker, Matthew J and Hussain, Syed R and Lovergne, Laurent and others},
  journal={Nature Protocols},
  volume={9},
  number={8},
  pages={1771--1791},
  year={2014}
}

@article{kong2015raman,
  title={Raman spectroscopy for medical diagnostics — From in-vitro biofluid assays to in-vivo cancer detection},
  author={Kong, Kevin and Kendall, Catherine and Stone, Nicholas and Notingher, Ioan},
  journal={Advanced Drug Delivery Reviews},
  volume={89},
  pages={121--134},
  year={2015}
}

@article{croce2014fluorescence,
  title={Fluorescence spectroscopy and imaging: basic principles and applications},
  author={Croce, Anna Chiara and Bottiroli, Giovanni},
  journal={European Journal of Histochemistry},
  volume={58},
  number={4},
  year={2014}
}

@article{cui2018machine,
  title={Machine learning and the physical sciences},
  author={Cui, Brandon and others},
  journal={Reviews in Physics},
  volume={3},
  pages={1--16},
  year={2018}
}

@article{shen2020deep,
  title={Deep learning in medical imaging and spectroscopy},
  author={Shen, Dinggang and Wu, Guorong and Suk, Heung-Il},
  journal={Annual Review of Biomedical Engineering},
  volume={19},
  pages={221--248},
  year={2020}
}

@article{hands2014brain,
  title={Brain tumour differentiation: rapid stratified serum diagnostics via attenuated total reflection Fourier-transform infrared spectroscopy},
  author={Hands, Jamie R and Abel, Paul and Ashton, Keith and others},
  journal={Journal of Neuro-Oncology},
  volume={127},
  number={3},
  pages={463--472},
  year={2016}
}

@article{paraskevaidi2018differentiating,
  title={Differentiating ovarian cancer from benign gynaecological conditions using Raman spectroscopy},
  author={Paraskevaidi, Maria and others},
  journal={Journal of Biophotonics},
  volume={11},
  number={3},
  year={2018}
}

@article{riosreina2019data,
  title={Data fusion approaches in spectroscopic analysis},
  author={R{\'\i}os-Reina, Rosa and others},
  journal={Talanta},
  volume={195},
  pages={560--572},
  year={2019}
}

@article{3d525b947d8b481e8ddfbc36f450e4fa,
title = "Characterizing dissolved organic matter fluorescence with parallel factor analysis: a tutorial",
author = {Colin A. Stedmon and Rasmus Bro},
year = "2008",
language = "English",
volume = "6",
pages = "572--579",
journal = "LIMNOLOGY AND OCEANOGRAPHY-METHODS",
issn = "1541-5856",
publisher = "John Wiley and Sons Inc.",
}

@article{bahram2006,
author = {Bahram, Morteza and Bro, Rasmus and Stedmon, Colin and Afkhami, Abbas},
title = {Handling of Rayleigh and Raman scatter for PARAFAC modeling of fluorescence data using interpolation},
journal = {Journal of Chemometrics},
volume = {20},
number = {3-4},
pages = {99-105},
doi = {https://doi.org/10.1002/cem.978},
url = {https://analyticalsciencejournals.onlinelibrary.wiley.com/doi/abs/10.1002/cem.978},
eprint = {https://analyticalsciencejournals.onlinelibrary.wiley.com/doi/pdf/10.1002/cem.978},
year = {2006}
}

@Article{C3AY41160E,
author ="Murphy, Kathleen R. and Stedmon, Colin A. and Graeber, Daniel and Bro, Rasmus",
title  ="Fluorescence spectroscopy and multi-way techniques. PARAFAC",
journal  ="Anal. Methods",
year  ="2013",
volume  ="5",
issue  ="23",
pages  ="6557-6566",
publisher  ="The Royal Society of Chemistry",
doi  ="10.1039/C3AY41160E",
url  ="http://dx.doi.org/10.1039/C3AY41160E"}

@article{https://doi.org/10.1002/cem.713,
author = {Shaffer, Ronald E.},
title = {Multi- and Megavariate Data Analysis. Principles and Applications, I. Eriksson, E. Johansson, N. Kettaneh-Wold and S. Wold, Umetrics Academy, Umeå, 2001, ISBN 91-973730-1-X, 533pp.},
journal = {Journal of Chemometrics},
volume = {16},
number = {5},
pages = {261-262},
doi = {https://doi.org/10.1002/cem.713},
url = {https://analyticalsciencejournals.onlinelibrary.wiley.com/doi/abs/10.1002/cem.713},
eprint = {https://analyticalsciencejournals.onlinelibrary.wiley.com/doi/pdf/10.1002/cem.713},
year = {2002}
}

@Article{ijms252010936,
AUTHOR = {Hano, Harun and Suarez, Beatriz and Lawrie, Charles H. and Seifert, Andreas},
TITLE = {Fusion of Raman and FTIR Spectroscopy Data Uncovers Physiological Changes Associated with Lung Cancer},
JOURNAL = {International Journal of Molecular Sciences},
VOLUME = {25},
YEAR = {2024},
NUMBER = {20},
ARTICLE-NUMBER = {10936},
URL = {https://www.mdpi.com/1422-0067/25/20/10936},
PubMedID = {39456720},
ISSN = {1422-0067},
DOI = {10.3390/ijms252010936}
}

@article{RIOSREINA2019560,
title = {Data fusion approaches in spectroscopic characterization and classification of PDO wine vinegars},
journal = {Talanta},
volume = {198},
pages = {560-572},
year = {2019},
issn = {0039-9140},
doi = {https://doi.org/10.1016/j.talanta.2019.01.100},
url = {https://www.sciencedirect.com/science/article/pii/S0039914019301201},
author = {Rocío Ríos-Reina and Raquel M. Callejón and Francesco Savorani and José M. Amigo and Marina Cocchi}
}

@article{CAMPOS2020103959,
title = {Data preprocessing for multiblock modelling – A systematization with new methods},
journal = {Chemometrics and Intelligent Laboratory Systems},
volume = {199},
pages = {103959},
year = {2020},
issn = {0169-7439},
doi = {https://doi.org/10.1016/j.chemolab.2020.103959},
url = {https://www.sciencedirect.com/science/article/pii/S0169743919305350},
author = {Maria P. Campos and Marco S. Reis}
}

@article{SILVESTRI2014181,
title = {A mid level data fusion strategy for the Varietal Classification of Lambrusco PDO wines},
journal = {Chemometrics and Intelligent Laboratory Systems},
volume = {137},
pages = {181-189},
year = {2014},
issn = {0169-7439},
doi = {https://doi.org/10.1016/j.chemolab.2014.06.012},
url = {https://www.sciencedirect.com/science/article/pii/S0169743914001336},
author = {M. Silvestri and A. Elia and D. Bertelli and E. Salvatore and C. Durante and M. {Li Vigni} and A. Marchetti and M. Cocchi}
}

@article{MISHRA2021116206,
title = {Recent trends in multi-block data analysis in chemometrics for multi-source data integration},
journal = {TrAC Trends in Analytical Chemistry},
volume = {137},
pages = {116206},
year = {2021},
issn = {0165-9936},
doi = {https://doi.org/10.1016/j.trac.2021.116206},
url = {https://www.sciencedirect.com/science/article/pii/S0165993621000285},
author = {Puneet Mishra and Jean-Michel Roger and Delphine Jouan-Rimbaud-Bouveresse and Alessandra Biancolillo and Federico Marini and Alison Nordon and Douglas N. Rutledge}
}

@article{KONG2015121,
title = {Raman spectroscopy for medical diagnostics — From in-vitro biofluid assays to in-vivo cancer detection},
journal = {Advanced Drug Delivery Reviews},
volume = {89},
pages = {121-134},
year = {2015},
note = {Pharmaceutical applications of Raman spectroscopy – from diagnosis to therapeutics},
issn = {0169-409X},
doi = {https://doi.org/10.1016/j.addr.2015.03.009},
url = {https://www.sciencedirect.com/science/article/pii/S0169409X15000447},
author = {Kenny Kong and Catherine Kendall and Nicholas Stone and Ioan Notingher}
}

@article{Shi,
	author = {Shi, Lingyan and Li, Yajuan and Li, Zhi},
	date = {2023/09/15},
	doi = {10.1038/s41377-023-01271-7},
	id = {Shi2023},
	isbn = {2047-7538},
	journal = {Light: Science \& Applications},
	number = {1},
	pages = {234},
	title = {Early cancer detection by SERS spectroscopy and machine learning},
	url = {https://doi.org/10.1038/s41377-023-01271-7},
	volume = {12},
	year = {2023}}

\end{document}